\documentclass[runningheads]{llncs}
\usepackage{times}
\usepackage{epsfig}
\usepackage{graphicx}

\usepackage{amsmath}
\usepackage{amssymb}

\usepackage{bm}
\usepackage{mathtools}

\usepackage{color}
\usepackage{booktabs}
\usepackage{multirow}
\usepackage{algorithm}
\usepackage[misc]{ifsym}

\DeclareMathOperator*{\argmin}{arg\,min}
\usepackage{subfig}               
\usepackage{overpic} 
\usepackage[compatible]{algpseudocode} 

\usepackage{algorithmicx,algorithm}

\begin{document}
%
\title{REST: Enhancing Group Robustness in DNNs through Reweighted Sparse Training}
\toctitle{REST: Enhancing Group Robustness in DNNs through Reweighted Sparse Training}

\author{Jiaxu Zhao \Letter\inst{1} \thanks{These authors contributed equally to this research. \\This paper was accepted by ECML PKDD 2023.} 
Lu Yin\inst{1}$^*$ 
Shiwei Liu\inst{2,1}  
Meng Fang\inst{3,1}
Mykola Pechenizkiy \inst{1} }
\authorrunning{J. Zhao et al.}
%
\institute{Eindhoven University of Technology, Eindhoven 5600 MB, Netherlands \\
\email{\{j.zhao,l.yin,s.liu,m.fang,m.pechenizkiy\}@tue.nl}\\ \and
The University of Texas at Austin, TX 78705 Austin, USA\\ 
\email{shiwei.liu@austin.utexas.edu}\\ \and
University of Liverpool, Liverpool L69 3BX, United Kingdom  \\
\email{Meng.Fang@liverpool.ac.uk}\\ } 
\tocauthor{Jiaxu Zhao, Lu Yin, Shiwei Liu, Meng Fang, Mykola Pechenizkiy}
\maketitle               

\begin{abstract}
The deep neural network (DNN) has been proven effective in various domains. However, they often struggle to perform well on certain minority groups during inference, despite showing strong performance on the majority of data groups.
This is because over-parameterized models learned \textit{bias attributes} from a large number of \textit{bias-aligned} training samples. These bias attributes are strongly spuriously correlated with the target variable, causing the models to be biased towards spurious correlations (i.e., \textit{bias-conflicting}). 
To tackle this issue, we propose a novel \textbf{re}weighted \textbf{s}parse \textbf{t}raining framework, dubbed as \textit{\textbf{REST}}, which aims to enhance the performance of biased data while improving computation and memory efficiency. 
Our proposed REST framework has been experimentally validated on three datasets, demonstrating its effectiveness in exploring unbiased subnetworks. We found that REST reduces the reliance on spuriously correlated features, leading to better performance across a wider range of data groups with fewer training and inference resources.
We highlight that the \textit{REST} framework represents a promising approach for improving the performance of DNNs on biased data, while simultaneously improving computation and memory efficiency. By reducing the reliance on spurious correlations, REST has the potential to enhance the robustness of DNNs and improve their generalization capabilities. Code is released at \url{https://github.com/zhao1402072392/REST}.

\end{abstract}

\section{Introduction}
Deep neural network models are successful on various tasks and can achieve low average error on the test set. However, these models may still exhibit high errors for certain subgroups of samples that are the minority in the training dataset \cite{duchi2019distributionally,zhao2023chbias}. This problem may be caused by the standard method (empirical risk minimization (ERM)) of training models that optimizes the average training loss.
During training, models learn the \textit{spurious correlations} that exist within the majority groups in the data and make predictions based on this correlation. 
For example, consider an image classification task of cows and camels, where most of the images of cows are captured in grasslands and camels in deserts. As a result of training on such a dataset, models tend to rely on features such as the presence of grass or sand in an image instead of the object of interest for making predictions \cite{beery2018recognition}. This can lead to poor performance of minority groups (e.g., cows in the desert and camels in the grassland). Since the model learns spuriously correlated features from the misleading statistical evidence in the data of the majority group, it will perform poorly on the minority group.

To address the problem that models perform poorly on minority groups and train an unbiased network that can perform well on biased datasets, researchers proposed various methods. One fundamental idea is to adjust the weights of different groups in the data during training. This approach, known as reweighting, has been extensively studied in the literature \cite{shimodaira2000improving,byrd2019effect,sagawa2020investigation}. The main goal of reweighting is to upweight the training loss for minority groups, thus encouraging the model to pay more attention to these groups and achieve higher accuracy on biased data.
Some recent studies have focused on improving the worst-group error, which measures the performance of the model on the subgroup with the lowest accuracy. One approach is based on distributionally robust optimization (DRO) \cite{duchi2016statistics,ben2013robust}, which aims to optimize the worst-case performance of the model under all possible distributions.  Based on distributionally robust optimization (DRO) \cite{duchi2016statistics,ben2013robust}, for instance, Sagawa et al.~\cite{sagawa2019distributionally} propose GDRO, a DRO-based method that directly minimizes the worst-group error during training. By doing so, the model can learn to avoid spurious correlations and focus on the most informative features for all groups.
\cite{nam2020learning,liu2021just} propose a two-step training approach. The idea is to first identify the data that the model predicts incorrectly after ERM training and upweight these data to train the model again. Despite the progress made by these methods, most existing work still focuses on training or fine-tuning model parameters in different ways to mitigate the bias of the model.

Some researchers found that over-parameterization allows the model to achieve a high average test accuracy but decreases the minority test accuracy due to capturing spurious correlations in the data \cite{sagawa2020investigation}.  Therefore, in addition to fine-tuning the model, some work \cite{zhang2021can} focused on improving the accuracy of the model on biased data by pruning the model parameters. For instance, Zhang et al.~\cite{zhang2021can} demonstrate that there exist sub-networks in the neural network that are less susceptible to spurious correlation. But all existing sparsity-based approaches rely on pruning a fully-trained dense network, which itself is very tedious. It is required to train a dense model first, then prune the model parameters according to some regulations to obtain a sparse model, and finally fine-tune the sparse model to recover accuracy. 


Overall, previous works address the model's poor performance on highly biased data either by using minority-group-aware optimization (e.g., loss reweighting, DRO \cite{duchi2016statistics,ben2013robust}, GDRO \cite{sagawa2019distributionally}), or by modifying the biased model using some ad hoc operations (e.g., pruning \cite{zhang2021can}, re-training on the biased data group (e.g., \cite{goel2020model}). In this paper, we propose to close this research question by directly training sparse neural networks from scratch to overcome the key hurdle of over-parameterization in memorizing spurious features. Our approach directly yields a sparse subnetwork that is debiased ``out of the box'', without any costly pre-training or any dense training steps.  
Specifically, we utilize sparse training \cite{mocanu2018scalable,liu2021we} to find the sparse subnetwork. Sparse training  has been proposed to solve the over-parameterization problem. It prunes the unimportant parameters of the model in training to sparse the over-parameterized model. And the model can retain the performance of the original model at a very low density. 

To the best of our knowledge, we use sparse training for the first time to solve the problem of high worst-group error in the model. By pruning certain parameters of the model during sparse training, we can create a sparse network that is less susceptible to spurious correlations and more robust to distribution shifts. We implement experiments on three popular image classification datasets (Colored MNIST (CMNIST) \cite{arjovsky2019invariant}, Corrupted CIFAR-10 (CIFAR-10-C) \cite{hendrycks2019benchmarking} and Gender-biased FFHQ (BFFHQ) \cite{kim2021biaswap}) to demonstrate the effectiveness of sparse training in optimizing the worst group errors. We also compare the performance of pruned pre-trained models and find that pruning the pre-trained models does not make them perform as well as sparse training models from scratch. Furthermore, we implement ablation experiments to compare the performance of different sparse training approaches in dealing with the out-of-distribution issue. We summarize our contributions as follows:

\begin{itemize}
    \item  \textbf{A New Approach:}  We propose the \textbf{Re}weighted \textbf{S}parse \textbf{T}raining (\textbf{REST}) framework, which trains a subnetwork in an end-to-end fashion. The framework uses sparse training to obtain a sub-network that avoids being biased towards spurious correlation in biased datasets and does not require an additional training and fine-tuning process.
    
    \item  \textbf{Better Performance:} We demonstrated that {\textbf{REST}} could achieve better performance than other strong baselines on all three highly biased datasets. Compared to the original ERM, with 0.5\% of bias-conflicting data, our method improves the accuracy by 28.3\%, 9.9\%, and 23.7\% on the CMNIST, CIFAR-10-C, and BFFHQ, respectively. 
    
    \item  \textbf{Fewer Resources:} Compared with other baselines, {\textbf{REST}} requires fewer training and inference resources. Specifically, after implementing our method, the ResNet-18 and Simple CNN models require only 2\% and 0.7\% of the original models' FLOPs for their application, respectively.

\end{itemize}

\section{Related Work}

\subsection{Sparse Neural Network Training}

Training sparse neural networks is a popular area of research. The goal is to train initial sparse networks from scratch and achieve comparable performance to dense networks while using fewer resources. Sparse training can be divided into two categories: static sparse training (SST), where the connectivity does not change during training, and dynamic sparse training (DST), where the connectivity changes during training.


\textbf{Static sparse training} refers to a set of techniques that involve training sparse neural networks while maintaining a consistent sparse connectivity pattern throughout the process. Despite the fixed sparse connectivity, there can be variations in layer-wise sparsity (i.e., the sparsity level of each individual layer). The simplest approach is to apply uniform sparsity to all layers \cite{gale2019state}. \cite{Mocanu2016xbm} introduced a non-uniform sparsity method that can be used in Restricted Boltzmann Machines (RBMs) and outperforms dense RBMs. Some research investigates the use of expander graphs for training sparse CNNs, demonstrating performance comparable to their dense counterparts \cite{prabhu2018deep,kepner2019radix}. Drawing from graph theory, the \textit{Erd{\H{o}}s-R{'e}nyi} (ER) model \cite{mocanu2018scalable} and its CNN variant, the \textit{Erd{\H{o}}s-R{'e}nyi-Kernel} (ERK) model \cite{evci2020rigging}, assign lower sparsity to smaller layers, thus preventing the layer collapse issue \cite{tanaka2020pruning} and generally yielding better results than uniform sparsity approaches. \cite{yin2022superposing,liu2021deep} combine individual subnetworks and  surpass the generalization performance of the naive dense ensemble. 


\textbf{Dynamic sparse training} involves training initial sparse neural networks while dynamically modifying the sparse connectivity pattern throughout the process. DST was first introduced by Sparse Evolutionary Training (SET) \cite{mocanu2018scalable}, which initializes sparse connectivity using an ER topology and periodically explores the parameter space via a prune-and-grow scheme during training. Subsequent to SET, weight redistribution has been introduced to search for optimal layer-wise sparsity ratios during training \cite {mostafa2019parameter,dettmers2019sparse}. The most commonly used pruning criterion employed in existing DST methods is magnitude pruning. Criteria for weight regrowth differ among methods, with gradient-based regrowth (e.g., momentum \cite{dettmers2019sparse} and gradient \cite{evci2020rigging}) demonstrating strong results in image classification, while random regrowth surpasses the former in language modeling \cite{dietrich2021towards}. 
Later research has improved accuracy by relaxing the constrained memory footprint \cite{jayakumar2020top,yuan2021mest,liu2021neuroregeneration,huang2022fedspa}.

\subsection{Debiasing Frameworks}
Neural networks tend to rely on spurious correlations in the data to predict, which are often caused by misleading statistical information in the data, and these spurious correlations do not generalize to all samples. Several works \cite{sagawa2020investigation,izmailov2022feature} have investigated the causes of worst-group errors as a result of models relying on spurious features of the data during training. These spurious features are relevant to the target but not to the research problem. 

To improve the performance of the network on biased data, a common approach is to reweight data from different distributions during training \cite{shimodaira2000improving,byrd2019effect,sagawa2020investigation}. \cite{sagawa2019distributionally} propose group distributional robust optimization DRO (GDRO) optimize the worst-group error directly. In addition to optimizing the performance of the worst group, some work \cite{khani2019maximum,agarwal2018reductions} attempt to improve the group robustness of the models by closing their performances in different groups.
\cite{yao2022improving,goel2020model} attempt to balance the training data through data augmentation techniques. \cite{geirhos2018imagenet} address the texture bias issue by incorporating additional training images with their styles transferred through adaptive instance normalization \cite{huang2017arbitrary}.

These methods above improve the performance of the models on biased data by using loss functions or adjusting the data distribution. There is also some work that considers the existence of unbiased sub-networks in the original network \cite{zhang2021can}. A common approach is to train a dense neural network first, then prune off some of the network weights to get a sparse network, and finally, fine-tune the sparse network. In this paper, we also focus on training a sparse network to improve its performance on biased data. However, instead of taking trivial steps, we train a sparse network directly from scratch through sparse training.


\section{Methodology}

Given a supervised dataset consisting of input samples $X \in \mathcal{X}$ and true labels $Y \in \mathcal{Y}$. We can denote the random input sample and its corresponding label as $(X^e, Y^e) \sim P^e$, where $X^e \in \mathcal{X}$ and $Y^e \in \mathcal{Y}$. Here, $e \in \mathcal{E} = \{1, 2, . . . E\}$ represents the index of the environment and $P^e$ represents the distribution associated with that environment. The set $\mathcal{E}$ contains all possible environments. Additionally, we assume that $\mathcal{E}$ is composed of training environments ($\mathcal{E}_{train}$) and unseen test environments ($\mathcal{E}_{test}$), such that $\mathcal{E}=\mathcal{E}_{train} \cup \mathcal{E}_{test}$. The training dataset comprises samples from $\mathcal{E}_{train}$. And the test dataset samples are from out-of-distribution in unseen environments $\mathcal{E}_{test}$.

Consider a neural network $f_\theta: \mathcal{X} \rightarrow \mathcal{Y}$ parameterized by $\theta$. Define the risk achieved by the model as $\mathcal{R}^e{(\theta)}=\mathbb{E}_{(X^e, Y^e)\sim P^e}[\ell(X^e, Y^e)]$, where $\ell$ is the loss of each sample (e.g.,cross-entropy). The objective of addressing the out-of-distribution (OOD) generalization problem is to develop a model that can effectively minimize the maximum risk across all environments in the set $\mathcal{E}$, as represented by the equation:

\begin{align}
  \min\limits_{\theta}\max\limits_{e\in \mathcal{E}} \mathcal{R}^e{(\theta)}
\end{align}
However, since we only have access to training data from $\mathcal{E}_{train}$ and cannot know samples from unseen environments, Therefore, the models tend to perform well on the data distribution on the training set but poorly on out-of-distribution data. 

Typically, models are trained in a way that optimizes the loss of model predictions and labels in the training environment, usually using empirical risk minimization (ERM). Usually, neural networks learn target features as well as spuriously correlated features, which are due to misleading statistical information. However, these spuriously correlated features do not generalize to all samples. Therefore, the models perform well in the training environment, and their predictions are highly accurate, but they perform poorly when they encounter samples without such spurious correlations. 

\begin{figure}[htbp]
	\centering
{\includegraphics[width=1\linewidth]{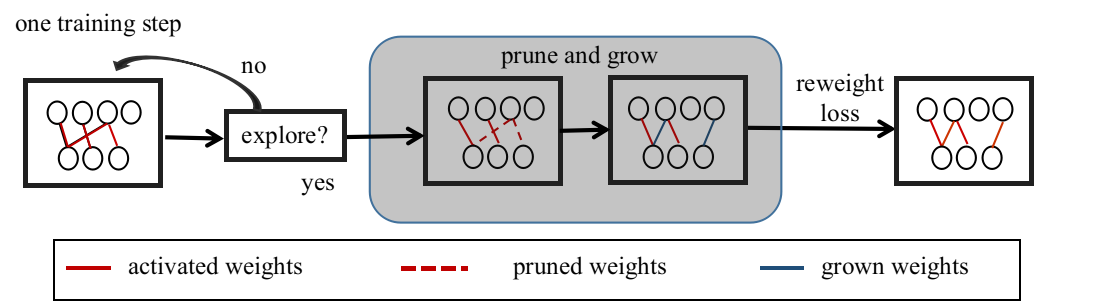}}

	\caption{Re-weighted Sparse Training.}
 \label{fig: sparse_structure}
\end{figure}

In this work, we proposed applying sparse training with reweighting to find a subnetwork to avoid learning spuriously correlated features. Our REST method is illustrated in Figure \ref{fig: sparse_structure}. In the following, we will formally demonstrate the details of sparse training.

\subsection{Sparse Training}

Let us denote the sparse neural network as $f(\bm{x}; \bm{\mathrm{\theta}}_\mathrm{s})$. $\bm{\mathrm{\theta}}_\mathrm{s}$ refers to a subset of the full network parameters $\bm{\mathrm{\theta}}$ at a sparsity level of ${(1 - \frac{\|\bm{\mathrm{\theta}}_\mathrm{s}\|_0}{\|\bm{\mathrm{\theta}}\|_0}})$ and $\|\cdot\|_0$ represents the $\ell_0$-norm. 

It is common to initialize sparse subnetworks $\bm{\mathrm{\theta}}_\mathrm{s}$ randomly based on the uniform \cite{mostafa2019parameter,dettmers2019sparse} or non-uniform layer-wise sparsity ratios with \textit{Erd{\H{o}}s-R{\'e}nyi} (ER) graph \cite{mocanu2018scalable,evci2020rigging}.  In the case of image classification, sparse training aims to optimize the following object using data $\{(x_i, y_i) \}_{i=1}^{\mathrm{N}}$:
\begin{equation}
\hat{\bm{\mathrm{\theta}}_\mathrm{s}} = \argmin_{\bm{\mathrm{\theta}}_\mathrm{s}} \sum_{i=1}^{\mathrm{N}} \mathcal{L}(f(x_i;\bm{\mathrm{\theta}}_\mathrm{s}),y_i)
\end{equation}
where $\mathcal{L}$ is the loss function.   

Static sparse training (SST) maintains the same sparse network connectivity during training after initialization. Dynamic sparse training (DST), on the contrary, allows the sparse subnetworks to  dynamically explore new parameters while sticking to a fixed sparsity budget. Most if not all DST methods follow a simple prune-and-grow scheme \cite{mocanu2018scalable} to perform parameter exploration, i.e., pruning $r$ proportion of the least important parameters based on their magnitude, and immediately grow the same number of parameters randomly \cite{mocanu2018scalable} or using the potential gradient \cite{liu2021we}. Formally, the parameter exploration can be formalized as the following two steps:
\begin{equation}
    \bm{\mathrm{\theta}}_{\mathrm{s}} = \Psi(\bm{\mathrm{\theta}}_\mathrm{s},~r),
    \label{eq:prune}
\end{equation}
\begin{equation}
    \bm{\mathrm{\theta}}_\mathrm{s} = \bm{\mathrm{\theta}}_{\mathrm{s}}
    \cup \Phi(\bm{\mathrm{\theta}}_{i \notin \bm{\mathrm{\theta}}_\mathrm{{s}}},~r)
    \label{eq:regrow}
\end{equation}

\noindent where $\Psi$ is the specific pruning criterion and $\Phi$ is growing scheme. These metrics may vary from one sparse training method to another. At the end of the training, sparse training can converge to a performant sparse subnetwork. Since the sparse neural networks are trained from scratch, the memory requirements and training/inference FLOPs are only a fraction of their dense counterparts.

It is well known that the standard method for training neural networks is Empirical Risk Minimization (ERM). Formally, the ERM can be defined as follows:
\begin{align}
\label{eq: erm}
  \theta_{ERM}=arg\min\limits_{\theta}\mathbb{E}_{(X^e, Y^e)\sim P^e}[\ell(X^e, Y^e)]
\end{align}
where $\ell$ is the cross-entropy loss or square loss. However, in previous works \cite{sagawa2019distributionally,sagawa2020investigation}, the authors demonstrate that models trained by ERM, whether under or over-parameterized, have low worst-group test errors (e.g., data that are not in the training set distribution). To address this issue, the reweighting method is the most common and simple method. Because the out-of-distribution problem is due to the tendency of the model to rely on strong spurious correlation in the training data to predict the results, the idea of reweighting is to reduce such spurious correlation by increasing the weight of minority groups in the data during training. Based on Equation \ref{eq: erm}, we formally define reweighting as the following equation:
\begin{align}
  \theta_{ERM}=arg\min\limits_{\theta}\mathbb{E}_{(X^e, Y^e)\sim P^e}[\beta_{P^e}\ell(X^e, Y^e)]
  \label{reweighted_func}
\end{align}
where $\beta_{P^e}$ is a reweighting hyperparameter. Usually, the $\beta_{P^e}$ is an upweight for a minority group of data in the training set and a downweight for a majority group. This can effectively mitigate the spurious correlations in the data learned by the network. Specifically, in some methods, $\beta_{P^e}$ is set according to the amount of data distribution (e.g., using $\frac{1}{N_{P^e}}$ as the $\beta_{P^e}$, where $N_{P^E}$ denotes the number of data from $P^e$ distribution.). In this paper, we set different $\beta_{P^e}$  for different training sets, which are described in Section \ref{sec: model setup}.







\section{Experiments}
In this chapter, we describe the details of the experiments. We conducted experiments on three datasets, including Colored MNIST (CMNIST) \cite{arjovsky2019invariant}, Corrupted CIFAR-10 (CIFAR-10-C) \cite{hendrycks2019benchmarking} and Gender-biased FFHQ (BFFHQ) \cite{kim2021biaswap}.
We choose a couple of debiasing methods as our baselines for comparison, including ERM, MRM \cite{zhang2021can}, and DisEnt \cite{lee2021learning}. Below, we describe the baseline methods, dataset, and model setup separately. 

\subsection{Baselines}

\paragraph{\textbf{ERM}}
Empirical Risk Minimization (ERM) is a technique used in machine learning to find the optimal parameters for a given model. ERM is the standard  baseline model in classification tasks. The basic idea behind ERM is to minimize the difference between the predicted output of a model and the ground truth label of the data. In addition to the original ERM, we also apply reweight loss to the ERM as another baseline. 

\paragraph{\textbf{MRM}}

Modular Risk Minimization (MRM) optimizes neural networks by learning a sparse subnetwork architecture that can improve the network's generalization performance. MRM consists of three stages, and we formally introduce MRM below:

Given data $(x_i, y_i)$, neural network $f(\mathbf{\theta};\cdot)$, subnetwork logits $\pi$ and the coefficient of sparsity penalty $\alpha$.

Stage 1: Full Model Pre-Training
In this stage, the full neural network model is trained using the cross-entropy loss (LCE) on the given dataset. The model parameters are initialized with $w_0$, and the optimization is performed for $N_1$ steps using gradient descent. The cross-entropy loss is defined as the sum of the logarithmic loss of each class label for each input in the training set. Formally, the $f$ can be updated through:
\begin{align}
  \mathcal{L}_{CE}(\theta):=\sum_i y_i logf(\theta;x_i)
\end{align}

Stage 2: Module Structure Probe
Subnetwork architecture is learned at this stage. The algorithm samples a binary mask vector $\mathbf{m}$ = sigmoid$\pi$, where $\pi$ is a learnable parameter that determines the importance of each weight in the network. The loss function in stage 2 is defined as:
\begin{align}
  \mathcal{L}_{MOD}(\theta)=\mathcal{L}_{CE}(\mathbf{m}\odot\mathbf{\theta})+ \alpha\sum_l \pi_l
\end{align}
where $l$ denotes the $i$-th layer of the network, $\alpha$ is a hyperparameter and $\odot$ refers to the Hadamard product.

Stage 3: Subnetwork Retrain
In this stage, the learned subnetwork architecture is used to retrain the full neural network model. The subnetwork mask vector $\mathbf{m}$ is obtained by applying hard thresholding $\mathbf{m}=\{\pi_l >0|l=1,2,\dots\}$. The model parameters are set back to their initial value $w_0$, and the optimization is performed for $N_1$ steps using $L_{CE}(\mathbf{m}\odot\mathbf{\theta})$.

The MRM algorithm iterates through Stages 2 and 3 until the desired level of sparsity is achieved in the learned subnetwork architecture. The resulting sparse network architecture can improve the generalization performance of the neural network by reducing overfitting and increasing its capacity to capture the relevant features of the input data.

\paragraph{\textbf{DisEnt}}
DisEnt \cite{lee2021learning} is a feature-level augmentation strategy. By utilizing additional synthesized biased features during the training, DisEnt performed well in classification and debiasing results. In DisEnt, two separate encoders are trained to embed images into two latent vectors corresponding to the target attributes and biased attributes, respectively. The network is then updated by concatenating the two hidden vectors and using them as inputs to the two classifiers. To further improve the process of learning the target feature vectors, swapping two potential vectors among the training sets is used to diversify the samples with conflicting biases. Specifically, DisEnt randomly permutes the target features and biased features in each mini-batch to obtain the swapped features.

While training with additional synthetic features helps to avoid spurious correlations, utilizing these features from the beginning of the training process does not improve denoising performance. More specifically, in DisEnt, feature augmentation is performed after some training iterations, when the two features are disentangled to a certain extent.

\subsection{Datasets}

\begin{figure}[!t]
    \centering
    \includegraphics[width=1\textwidth]{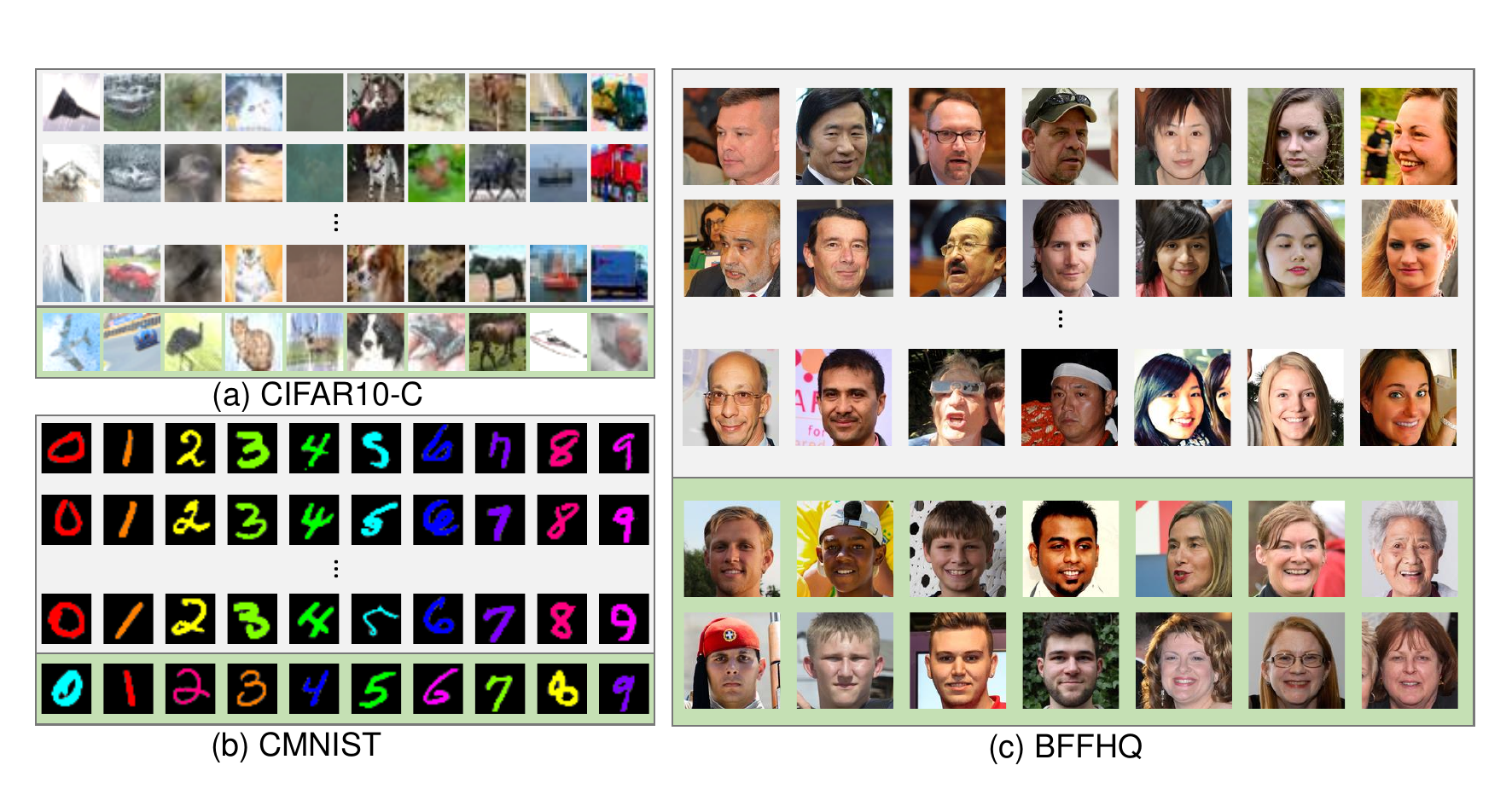}
    \caption{Examples of datasets. Sub-figure (a) is samples from the Colored MNIST dataset, sub-figure (b) is samples from the Corrupted CIFAR-10 dataset, and sub-figure (c) is samples from the Gender-biased FFHQ dataset. Each dataset consists of two more parts of data: the data at the upper part (gray background box) is the bias-aligned majority data set, and the data at the lower part (green background box) is the bias-conflicting minority data set.}
    \label{fig: dataset}
\end{figure}

We perform image classification experiments on three popular image datasets. As shown in Figure \ref{fig: dataset}, most of the data in the dataset are bias-aligned (gray background box), while a few are bias-conflicting (green background box). For Colored MNIST (CMNIST) and Corrupted CIFAR-10 (CIFAR-10-C), according to the different proportions of bias-conflicting samples ( 0.5\%, 1\%, 2\%, and 5\%), we have four train sets. For Gender-biased FFHQ (BFFHQ), the bias-conflicting data is 0.5\%.

\paragraph{\textbf{Colored MNIST (CMNIST)}}
Colored MNIST (CMNIST) \cite{arjovsky2019invariant} was proposed to be a variant of the original MNIST dataset \cite{lecun1998gradient}, In Colored MNIST, color information is added to the images by assigning a random RGB color value to each pixel in the image. Therefore, unlike the original MNIST dataset, CMNIST has a label indicating the color in addition to a target label. We use the number corresponding to the image as the target label and the color as the bias label. Inspired by \cite{zhang2021can,lee2021learning}, we chose the dataset with ten different colors and set a one-to-one correspondence with the ten numerical labels (e.g., ``2''$\leftrightarrow$ ``green'', ``4'' $\leftrightarrow$``yellow''). Images that did not follow this correspondence are randomly assigned colors. We name the data that follow this correspondence \textit{bias-align} and the data that do not follow this correspondence \textit{bias-conflict}. So our dataset consists of \textit{bias-align} and \textit{bias-conflict}. The training data has the majority of \textit{bias-align} data (e.g., 99$\%$). The data in the test set do not have the number label and color label correlation that exist in the training set. So the test set can be used to evaluate the out-of-distribution performance of the model. 

\paragraph{\textbf{Corrupted CIFAR-10 (CIFAR-10-C)}}

Corrupted CIFAR-10 (CIFAR-10-C) \cite{hendrycks2019benchmarking} is a variant of CIFAR-10 dataset. a dataset is obtained by applying ten textures to the images of the b dataset. In the CIFAR-10-C, the classes of images and the types of textures applied are highly correlated, i.e., most of the images in the same class have the same texture applied to them. So the model trained on a dataset with such a strong association will perform poorly when tested on a data distribution without that association. In our experiments, CIFAR-10-C also consists of \textit{bias-conflict} and \textit{bias-align}. Similar to the CMNIST, we also constructed four training sets based on the different proportions of \textit{bias-conflict} samples.

\paragraph{\textbf{Gender-biased FFHQ (BFFHQ)}}
Gender-biased FFHQ (BFFHQ) \cite{kim2021biaswap} is built from the data of FlickrFaces-HQ (FFHQ) \cite{karras2019style}. FFHQ contains face images with multiple face attribute labels, including age, race, gender, glasses and hat. We choose ``age'' and ``gender'' as the target and bias labels. In the training set, most of the images have a ``young'' attribute for ``female'' and an ``old attribute for ``male''. So the ``younger'' label in the training set has a strong correlation with the ``female'' attributes, which will lead the model to classify the age of the images based on the gender attributes. We use a training set in which the sample of \textit{bias-conflict} is 0.5\%.

\subsection{Setup}
\label{sec: model setup}

\paragraph{Models}
For CMNIST, we use a convolutional network with three convolution layers, in which the feature map dimensions are 64, 128, and 256, respectively (named ``Simple CNN'' in this paper). Following \cite{zhang2021can,park2022efficient}, we add ReLU activation and a batch normalization layer after each convolution layer. For CIFAR-10-C and BFFHQ, we use ResNet-18 \cite{he2016deep}. 

\paragraph{Training details}
For the reweighting parameter $\beta_{P^e}$ in the reweighting, we set as $\{10, 30, 50, 80\}$ for CMNIST and CIFAR-10-C with\{0.5\%, 1.0\%, 2.0\%, 5.0\%\} of bias ratio, respectively. For BFFHQ, we set $\beta_{P^e}=$80. We use the Adam optimizer \cite{kingma2014adam}, weight decay = $1\cdot {10}^{-4}$, $\beta1$ = 0.9, $\beta2$ = 0.999. We use the learning rate$=1\cdot {10}^{-2}$ for CMNIST and $1\cdot {10}^{-3}$ for CMNIST and BFFHQ. We set the $update\ frequency =1000$ as the number of iterations to train between parameter explorations. We set up three different seeds and report their average values as the experimental results.





\begin{table}[!t]
\centering  
\caption{The results of floating-point operations per second (FLOPs) during training/testing and the number of parameters. We report the results of CIFAR-10-C for ResNet-18 and BFFHQ for Simple CNN.}
\begin{tabular}{l|ccc||ccc} 
        \toprule
         ~&\multicolumn{3}{c||}{{\textbf{ResNet-18}}} & \multicolumn{3}{c}{{\textbf{Simple CNN}}}\\ 
         \midrule
         {\textbf{Models}}
         & FLOPs &FLOPs& Para & FLOPs& FLOPs & Para \\
       
        \midrule
        Dense& $1\times(4.41e16)$&$1\times(3.27e9)$ &11.17M&$1\times(5.68e16)$ &$1\times(3.44e9)$ &0.661M\\
        \midrule
        REST& $0.02\times$& $0.016\times$&0.056M& $0.007\times$& $0.0075\times$&0.0032M\\
        \bottomrule
    \end{tabular}
    \label{tab: FLOPs}
\end{table}

\begin{table}[!t]
\centering 
\caption{The accuracy of image classification on three datasets. We conduct experiments on a simple CNN and model ResNet-18. For Colored MNIST and Corrupted CIFAR-10, we evaluate the accuracy using unbiased test sets and report the unbiased test accuracy. For BFFHQ, we report the bias-conflicting test accuracy. ``Ratio (\%)'' denotes the proportion of bias-conflict data in the training data. For our proposed method, we report the accuracy of REST under the best density (the numbers in parentheses). The best-performing results are indicated in bold.}
    \begin{tabular}{lcccccc} 
        \toprule 
        \textbf{Dataset}  & \textbf{Model}& \textbf{Ratio (\%)} &\textbf{ERM} & \textbf{MRM} & \textbf{DisEnt}  & \textbf{REST} \\
        \midrule
        \multirow{4}*{\textbf{CMNIST}}
         & \multirow{4}*{\textbf{Simple CNN}}~ & 0.5 & 45.7 & 43.5 & 43.1  & \textbf{48.3} (0.050)\\
        ~ &~ & 1.0 & 69.6& 70.3 & 65.9 & \textbf{72.1} (0.050) \\
        ~ &~ & 2.0 & 83.3& 83.9 & 79.8  & \textbf{84.8} (0.005) \\
        ~ &~ & 5.0 & 93.2& 92.8 & 92.3  & \textbf{93.7} (0.005) \\
        \midrule
        \multirow{4}*{\textbf{CIFAR-10-C}} & \multirow{4}*{\textbf{ResNet-18}}
         ~ & 0. & 25.9& \textbf{26.7} & 18.5  & \textbf{26.7} (0.050)\\
        ~ & ~ & 1.0 & 27.4& 26.8 & 21.0 & \textbf{28.8} (0.005) \\
        ~ & ~ & 2.0 & 31.0& 30.1 & 25.9 & \textbf{33.1} (0.005) \\
        ~ & ~ & 5.0 & 37.0& 37.7 & 39.2 & \textbf{39.4} (0.005) \\
        \midrule
        \textbf{BFFHQ}& \textbf{ResNet-18}& 0.5 & 41.8 & 53.5 & 54.2 & \textbf{63.5} (0.0005) \\
        \bottomrule
    \end{tabular}
\label{tab: baseline}
\end{table}

\subsection{Computational Costs}
In order to understand the computational demands of various methods, we utilized FLOPs as a measure of the computational consumption of each method. FLOPs stands for Floating Point Operations, and it is a metric used to quantify the number of arithmetic operations (addition, subtraction, multiplication, and division) carried out by a computer when executing a particular algorithm or method.
To further explore the computational consumption of different methods, we evaluate their performance during training and testing of both ResNet-18 and convolutional neural networks. 
During the evaluation process, we computed the number of FLOPs required for each method to complete a task or operation within the networks. By comparing the number of FLOPs required by each method, we were able to assess their relative efficiency and identify the methods that consume less computational resources. 

We report the results of FLOPs in Table \ref{tab: FLOPs}. For ResNet-18, the dense model required high FLOPs during training at $4.41e16$, which is significantly higher than our method. During testing, the dense model also required high FLOPs, with a value of $3.27e9$. Our method only requires 2\% and 1.6\% FLOPs of the dense model for training and testing, respectively. And our method only retains 0.49\% and 0.48\% number of the parameters of ResNet-18 and Simple CNN, respectively.

\subsection{Main Results}
We analyze the experimental results of our proposed method and compare it with three baseline methods: ERM, MRM \cite{zhang2021can}, and DisEnt \cite{lee2021learning}. We evaluate the performance of these methods on three different datasets: Colored MNIST, Corrupted CIFAR-10, and BFFHQ.

From Table \ref{tab: baseline}, we can see that our proposed method outperforms all the baseline methods on all three datasets. In particular, for the BFFHQ dataset, our method demonstrates a significant improvement over the original ERM, with a large gap. This indicates that our method is effective in handling bias-conflicting data and improving the accuracy of image classification.

It is also notable that the sparsities with the best performance during sparse training are very high, retaining only 0.05, 0.005, and even 0.0005 of the parameters of the original neural network. This suggests that our method is capable of achieving good performance with highly compact neural networks, which can be beneficial for real-world applications where computational resources are limited.

In terms of the baseline methods, we can see that MRM and DisEnt generally perform better than the original ERM on all datasets, but they are still outperformed by our proposed method. This highlights the effectiveness of our approach in handling bias-conflicting data and improving the robustness of image classification.

Our experimental results in Table \ref{tab: baseline} demonstrate that our proposed method is effective in handling bias-conflicting data and improving the accuracy of image classification. Moreover, it shows that highly compact neural networks can achieve good performance, which can be beneficial for real-world applications with limited computational resources.

\begin{table}[!t]
\centering

\caption{Performance of different sparse training on three data. We report the best performance of each sparse training method among the sparsity set. The number in parentheses indicates the sparsity level. The best results are indicated in bold.}
    \begin{tabular}{lcccccr} 
        \toprule 
        \textbf{Dataset}&\textbf{Model}&\textbf{Ratio(\%)} & \textbf{RigL}  & \textbf{REST}  \\
        \midrule
        \multirow{4}*{{\textbf{CMNIST}}}&\multirow{4}*{{\textbf{Simple CNN}}}
        ~ & 0.5  & 36.3 (0.050)  & \textbf{48.3} (0.050) \\
        ~ &~ & 1.0 & 60.1 (0.050) & \textbf{72.1} (0.050) \\
        ~ &~ & 2.0 & 74.7 (0.005)  & \textbf{84.8} (0.005)  \\
        ~ &~ & 5.0  & 89.5 (0.005) & \textbf{93.7} (0.005) \\
        \midrule
        \multirow{4}*{{\textbf{CIFAR-10-C}}}&\multirow{4}*{{\textbf{ResNet-18}}}
        ~ & 0.5  & 23.5 (0.050) & \textbf{26.7} (0.050)\\
        ~ &~ & 1.0 & 26.0 (0.005) & \textbf{28.8} (0.005)\\
        ~ &~ & 2.0 & 30.9 (0.005) & \textbf{33.1} (0.005)\\
        ~ &~ & 5.0  & 37.4 (0.005) & \textbf{39.4} (0.005)\\
        \midrule
        \textbf{BFFHQ}& \textbf{ResNet-18}&0.5 & 50 (0.0005) & \textbf{63.6} (0.0005)\\
       
        \bottomrule
    \end{tabular}
	
    \label{tab: reweight}
\end{table}

 

\begin{figure}[h]
	\centering
	\subfloat[Ratio 0.5\%]{\includegraphics[width=.48\linewidth]{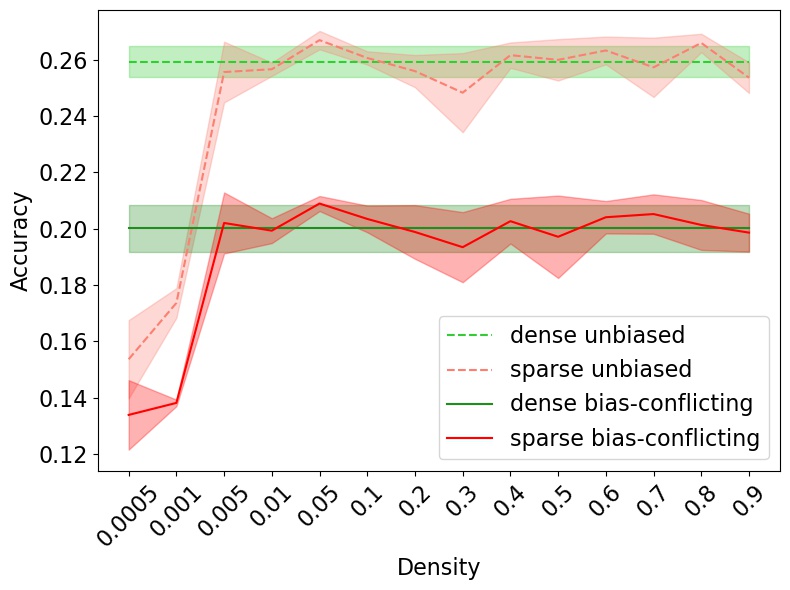}}\hspace{5pt}
	\subfloat[Ratio 1\%]{\includegraphics[width=.48\linewidth]{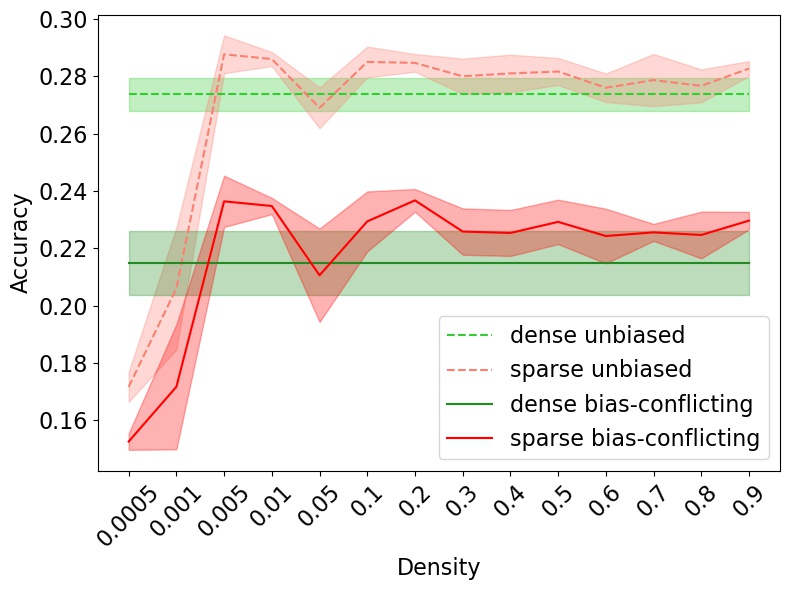}}\\
	\subfloat[Ratio 2\%]{\includegraphics[width=.48\linewidth]{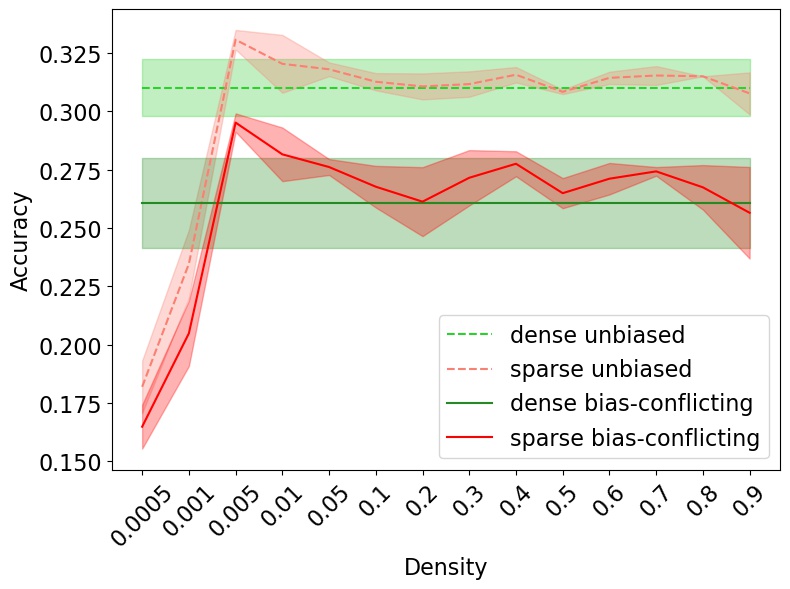}}
	\subfloat[Ratio 5\%]{\includegraphics[width=.48\linewidth]{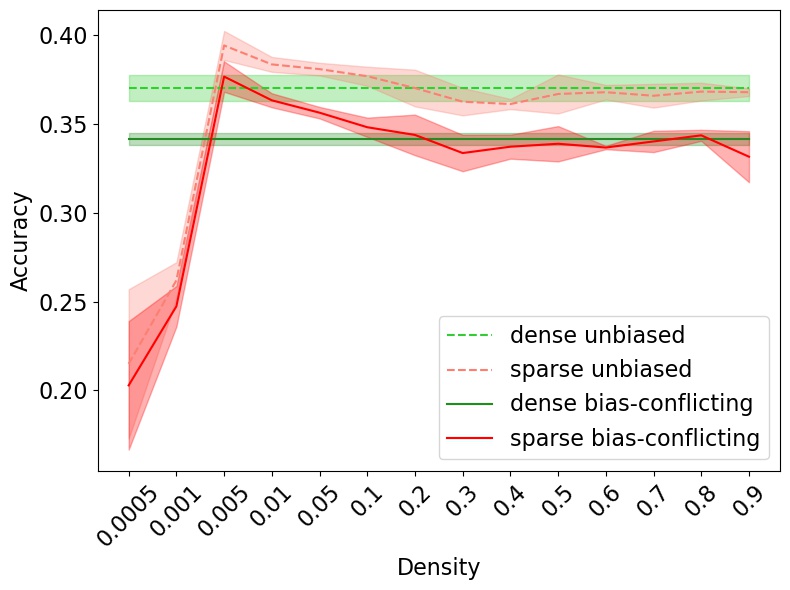}}
	\caption{Performance of RigL under different sparsity on CIFAR-10-C datasets with different ratios of bias-conflicting data. The green line represents the performance of the dense model, and the red line represents the performance of the sparse model after different sparse training. The solid line indicates the accuracy of bias-conflicting data in the test data, and the dashed line indicates the unbiased test accuracy.}
    \label{fig: ratio}
\end{figure}

\subsection{Ablation Study}
We also implemented an ablation experiment to help analyze the effectiveness of our method. We report the performance difference between RigL and our method on three datasets in Table \ref{tab: reweight}. Also, we compare the performance of our method at different sparsity levels. Specifically, we report the varying sparsity levels on the CIFAR-10-C dataset with 5\% bias-conflicting data. The CIFAR-10-C dataset consisted of images of objects from ten different classes, and it contains a diverse set of image corruptions.

\paragraph{\textbf{Diverse Sparse Levels}}
In this ablation study, we investigated the impact of sparsity on the performance of sparse models. To achieve this, we conducted experiments at fourteen different densities ranging from 0.0005 to 0.9 and compared the performance of sparse models to that of a dense model.

As depicted in Fig \ref{fig: ratio}, the performance of the sparse models were evaluated based on their image classification accuracy, and the red line represented the accuracy of the sparse model after our methods were applied. We also evaluated the accuracy of both bias-conflicting and unbiased test data. It was observed that the accuracy of the unbiased test was consistently higher than that of the bias-conflicting data for both models, indicating the problem of spurious correlation.

Furthermore, the level of density had a significant impact on the accuracy of the model. When the density was too low, the model's accuracy was very low, and when the density was increased to 0.005, the model's accuracy improved significantly. This indicates that some density levels can help the model avoid learning spuriously correlated features. However, as the density increased beyond this point, the accuracy of the model gradually decreased, indicating that the model was learning more spuriously correlated features as the number of model parameters increased.

The results in \ref{fig: ratio} highlight the effectiveness of our method that balances model complexity and performance by avoiding over-parameterization. It also emphasizes the importance of selecting an appropriate density level to optimize the performance of sparse models. Overall, our findings from Fig \ref{fig: ratio} provide valuable insights into the impact of sparsity on the performance of sparse models, which can help develop more efficient and accurate models.


\paragraph{\textbf{Diverse Bias-conflicting Data Ratio}}
In addition to evaluating the impact of sparsity on the performance of sparse models, Fig \ref{fig: ratio} also presents the performance of sparse training on data with varying bias-conflicting ratios. As shown in Fig \ref{fig: ratio}, the accuracy of both dense and sparse models increases with a higher proportion of bias-conflicted data. This suggests that models trained on bias-conflicting data are more robust and have a better generalization performance.

Moreover, it was observed that sparse training effectively enhances the model's performance across diverse bias-conflicting ratio data. This indicates that our method can improve the model's ability to generalize to various types of data with different levels of bias. By leveraging sparsity regularization, the model can learn to identify the most important and relevant features of the data while ignoring irrelevant and spurious features, leading to improved performance and generalization.

Overall, these findings further demonstrate the effectiveness of our method in enhancing the performance of out-of-distribution generalization, even in the presence of varying levels of bias-conflicting data. Our results suggest that applying our method in the training process can help to improve the robustness and generalization performance of machine learning models, which can be beneficial in various real-world applications.

\section{Conclusion}
The over-parameterized models learned from a large number of bias-aligned training samples often struggle to perform well on certain minority groups during inference, despite showing strong performance on the majority of data groups. To address this issue, we proposed a novel reweighted sparse training framework called REST, which aims to enhance the performance on various biased datasets while improving computation and memory efficiency, through the lens of sparse neural network training. Our experimental results demonstrate that REST can reduce the reliance on spuriously correlated features and improve performance across a wider range of data groups with fewer training and inference resources. By reducing the reliance on spurious correlations, REST represents a promising approach for improving the performance of DNNs on biased datasets while simultaneously improving their robustness and generalization capabilities.


\section*{Acknowledgements}
This work used the Dutch national e-infrastructure with the support of the SURF Cooperative
using grant no. EINF-3953/L1.

\newpage

\section*{Ethical Statement}
As researchers in the field of deep neural networks, we recognize the importance of developing methods that improve the generalization capabilities of these models, particularly for minority groups that may be underrepresented in training data. Our proposed reweighted sparse training framework, REST, aims to tackle the issue of bias-conflicting correlations in DNNs by reducing reliance on spurious correlations. We believe that this work has the potential to enhance the robustness of DNNs and improve their performance on out-of-distribution samples, which may have significant implications for various applications such as healthcare and criminal justice. However, we acknowledge that there may be ethical considerations associated with the development and deployment of machine learning algorithms, particularly those that may impact human lives. As such, we encourage the responsible use and evaluation of our proposed framework to ensure that it aligns with ethical standards and does not perpetuate biases or harm vulnerable populations.

\clearpage
\newpage
{\small
\bibliographystyle{splncs04}
\bibliography{ref}
}
\end{document}